\documentclass[letterpaper, 10 pt, conference]{ieeeconf}  % Comment this line out if you need a4paper
\IEEEoverridecommandlockouts    % This command is only needed if 
\usepackage{times}
\usepackage{epsfig}
\usepackage{graphicx}
\usepackage{subcaption}
\usepackage{amsmath,amssymb,amsfonts}
\usepackage{textcomp}
\usepackage{xcolor}
\usepackage{stackengine}
\usepackage{color,soul}
\usepackage{algorithm}
\usepackage{algpseudocode}
\usepackage{hyperref}
\DeclareMathOperator*{\argmax}{arg\,max}

\begin{document}

%%%%%%%%% TITLE
%\title{\LARGE \bf What we see and What we don't see:\\ a Robotics Mystery in Occluded Crowd Prediction}
\title{\LARGE \bf What we see and What we don't see:\\ Imputing Occluded Crowd Structures from Robot Sensing}
\author{Javad Amirian\thanks{The research is supported by the CrowdBot H2020 EU Project
		\url{http://crowdbot.eu/}}\\
	Univ Rennes, Inria, CNRS, IRISA\\
	France\\
	{\tt\small javad.amirian@inria.fr}
	\and
	Jean-Bernard Hayet\thanks{J.B. Hayet is partially funded by the Intel Probabilistic Computing initiative.}\\
	CIMAT, A.C.\\
	M\'exico\\
	{\tt\small jbhayet@cimat.mx}
	\and
	Julien Pettr{\'e}\\
	Univ Rennes, Inria, CNRS, IRISA\\
	France\\
	{\tt\small julien.pettre@inria.fr}
}

\maketitle
%\thispagestyle{empty}

%%%%%%%%% ABSTRACT
\begin{abstract}

We consider the navigation of mobile robots in crowded environments, for which onboard sensing of the crowd is typically limited by occlusions. We address the problem of inferring the human occupancy in the space around the robot, in blind spots, beyond the range of its sensing capabilities. This problem is rather unexplored in spite of the important impact it has on the robot crowd navigation efficiency and safety, which requires the estimation and the prediction of the crowd state around it. In this work, we propose the first solution to sample predictions of possible human presence based on the state of a fewer set of sensed people around the robot as well as previous observations of the crowd activity.

\end{abstract}

%\input{test}

% ====================
% !TeX root=main.tex

\section{Introduction}
\label{sec:Intro}
As French economist Fr\'ed\'eric Bastiat pointed out back in the 19th century, to make wise decisions in the economy, ``it is important to consider, in addition to what we see, what we do not see''. We believe such an elementary rule that once has been neglected in some economic decisions, is missing from one of the most critical tasks of robots: navigation in crowded environments, where the on-board sensing of the crowd is typically limited by occlusions and can substantially impact the robot's ability to anticipate possible collisions with occluded agents.
% from these blind spots.

With a growing presence of robots in public environments, situations in which we have to share our space with them are more and more common. In this context, and as claimed by other authors, a more accurate prediction of the near-future evolution of the environment helps the robots to reduce their re-planning effort~\cite{pfeiffer2016predicting}. That is why the topics of human trajectory prediction has become very important towards the goal of harmonious co-existence. However, an equally important topic is crowd state imputation (i.e., filling out the missing data in the occupancy map around the robot) to tackle potentially unobserved surrounding agents and it has not received the same attention.

In this paper, we address the problem of 
crowd state imputation from a mobile robot perspective.
When dealing with this problem from the robot perspective, the input data have properties that makes the prediction much harder, e.g. as opposed to more classical motion prediction problem with data coming from a static, bird-view surveillance camera. In particular, due to the low height of the sensor (e.g. LiDAR) installed on the robot, noticeable parts of the crowd can be occluded. Moreover, many pedestrians may remain undetected for long sequences of frames. Depending on the density of the crowd and the characteristics of the robot's sensor, the proportion of non-detected people can be negligible or severe.

\begin{figure}
	\centering
	\includegraphics[trim=80 40 80 40, clip, width=1.0\linewidth]{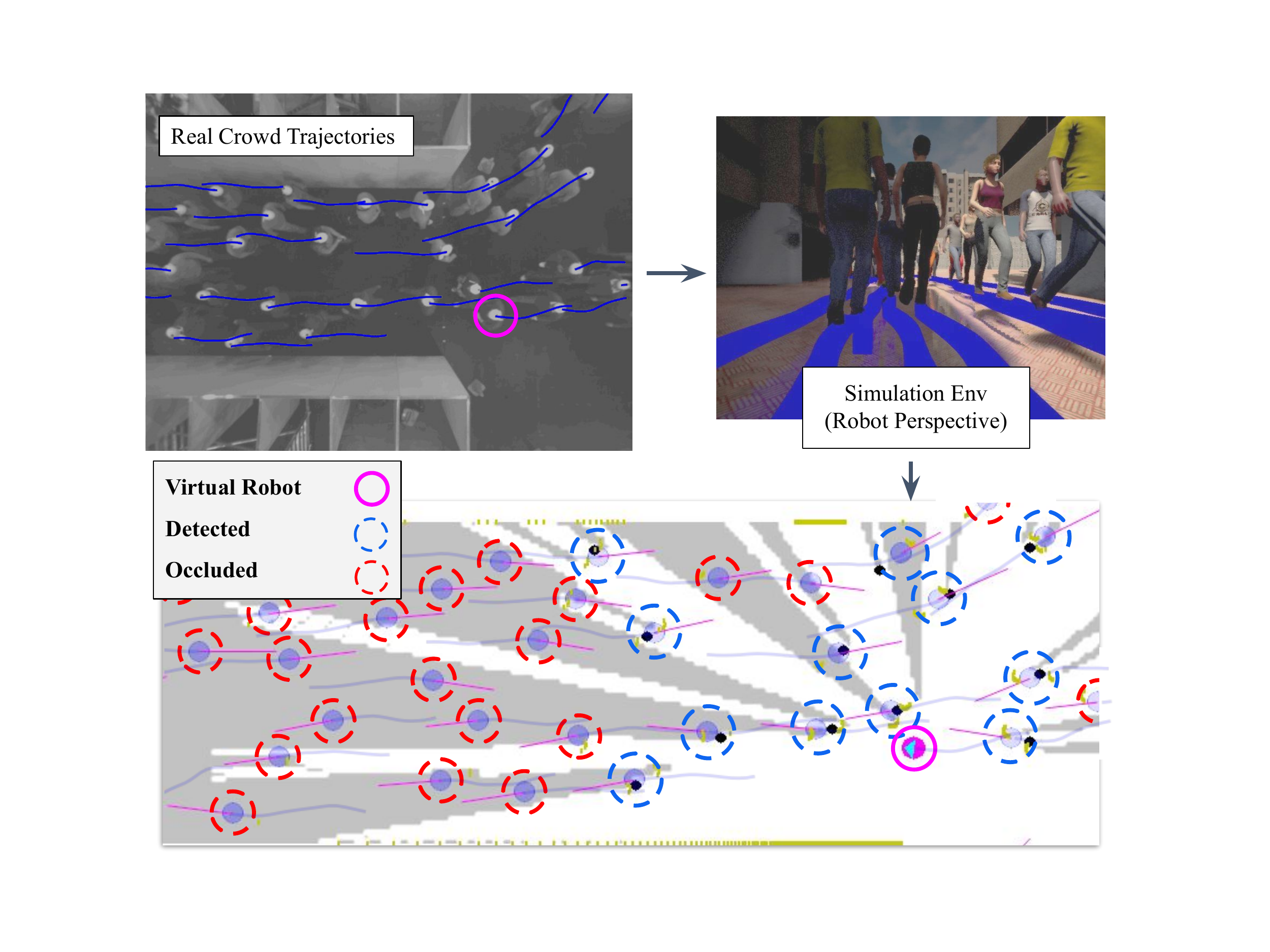}
	\caption{Occluded crowd: In the top left, a top-view image of a crowd is shown. The pedestrians form a bidirectional flow in a narrow corridor. If we \textit{simulate} a robot through one of the pedestrians (simulation of robot perspective in the top right image), there would be noticeable occlusion (gray area at the bottom) where the pedestrians are un-detected (red circles). The black dots within the blue circles show the detection locations, before filtering. Note: the crowd activity is recorded in the form of xy trajectories (blue lines).}
	\label{fig:occluded-problem}
\end{figure}

This can impact the performance of the robot in predicting future collisions and building a safe and valid motion plan. An example of typical crowd perception from a mobile robot, with multiple occlusions, is presented in Fig.~\ref{fig:occluded-problem} for a simulated mobile robot within a high-density crowd.

Our proposal is to leverage the statistical patterns extracted from past observations over a surrounding crowd to estimate the probability of the presence of people in unseen areas, i.e. we perform statistical imputation of the occupancy levels in these areas. Our model takes as input the stream of range data and the positions of the detected persons and gives as an output a prediction of the surrounding crowd motion. We state the problem as follows:
``Given the robot observations about the crowd surrounding it and given a query point $\textbf{x}$ (potentially out of robot's sight), what is the probability of presence of a person, that is not already detected, at $\textbf{x}$?''.

In most related research, if no person is detected/tracked at an unobserved location $\textbf{x}$, then no mechanism allows anticipating a potential collision coming from an agent located there. Addressing this problem is the major contribution of our work. To the best of our knowledge, this is the first effort to use the statistical crowd patterns in order to predict the state of people around a robot.

This paper is structured as follows:  We review related work in section~\ref{sec:related-work}. Then we formalize the problem and elaborate the details of the proposed method in section~\ref{sec:proposed}.
In section~\ref{sec:experiments}, we present the evaluation methodology and experimental results on real and simulated crowd data.
Finally, we conclude and discuss the future works in section~\ref{sec:conclusion}.

%\begin{figure}
%	\centering
%	\includegraphics[trim=20 0 20 0, clip, width=0.9\linewidth]{figs/Followbot-Motiv}
%	\caption{Partially occluded crowd. In the left picture, a robot (green circle) is seen moving in a crowd flow. The persons detected by the LiDAR (in blue) generate blind spots that contain other persons (in gray) not detected by robot. In the right picture, using our crowd synthesis algorithm, a few possible synthetic people are added that will impact on the prediction of other agents. 
%	}
%	\label{fig:occluded-crowd}
%\end{figure}

% !TeX root=main.tex

\section{Related Work}
\label{sec:related-work}

%Our work relies upon one main assumption: based on partial observations over the surrounding environment, an agent can make plausible hypothesis about the unseen static and dynamic obstacles (i.e., the environment is homogeneous).

%\subsection{
%	Prediction of Static Unknown Spaces
%	}

Prediction is one of the critical tasks of robots that help them deal with limited observation both in \textit{time} and \textit{space} domains.
Prediction of static unknown spaces takes inspiration in neuroscience research. Cognitive psychologists have suggested that humans are able to explore new unknown environments by making predictions of the occupied space beyond their current line of sight~\cite{buckner2010role}. The authors of~\cite{katyal2019uncertainty} study the ability to generate future predictions of occupancy maps using a U-Net style AutoEncoder neural network.
Wang et al.~\cite{wang2020learning} have proposed a neural-network-based method to predict the occupancy of the unknown space. Their model utilizes contextual information about the environment and learns from prior knowledge to predict the obstacles distribution in occluded spaces. The FASTER model~\cite{tordesillas2020faster} has shown that optimizing the robot local planner by considering both the known and unknown free spaces, can lead to higher-speeds and safer maneuvers for UAVs and ground robots.

%\subsection{
%	Prediction of Crowd Motion
%	}

Crowd Motion Prediction research focuses more on prediction on time domain, and is becoming a key building block for social robot navigation and self-driving vehicles. Social-LSTM~\cite{SocialLSTM2016} predicts the joint motion of dynamic pedestrians in crowded scenes by using LSTM networks and by pooling hidden states of neighboring agents. 
Social-GAN~\cite{SocialGAN2018} and Social-Ways~\cite{SocialWays2019} were proposed later to cope with multi-modal predictive distributions of future trajectories, using Generative Adversarial Networks.
In~\cite{pfeiffer2018datadriven}, both static obstacles and surrounding pedestrians are used for trajectory forecasting.
In the above works, the prediction is performed under assumption of a fully-observable environment.

Other works have addressed the trajectory forecasting problem with first-person- (or robot-) view perception to deal with occlusions. The authors of~\cite{bi2020FvTraj} have created a simulation environment using Unity game engine, and have simulated the view of pedestrians in a crowd for prediction of trajectories. In~\cite{pfeiffer2016predicting}, an interaction-aware motion model is learned based on human-human interactions observed by the robot with onboard sensors, but it is experimented only in very low-density scenarios.

The idea of extracting crowd patterns from collective behaviors has a long history. For example, Moussaid et. al~\cite{moussaid2010walking} have studied the grouping behavior on crowd videos recorded in public places and have reported the patterns such as group sizes and the distance and angle between the group members. In~\cite{dupont2017crowd11}, the authors propose a classification of crowd structures inspired from fluid dynamics. A common mathematical framework to study these crowd structures is probabilistic graph models, and in particular Conditional Random Fields~\cite{pellegrini2010improving}. Alahi et. al \cite{alahi2014SAM} have created a huge dataset of human trajectories, recorded in a train station, and have proposed \textit{Social Affinity Maps (SAM)} to capture the spatial position of an agent’s neighbors by radially binning the positions of his neighboring agents. In fact, they end up with a 10-bit binary radial histogram which is suggested to be a robust feature, not changing frequently and significantly over time. This feature improves the re-identification of groups in consequent \textit{tracklet}s extracted from different cameras in the train station.

% !TeX root=main.tex

\section{Proposed Method}
\label{sec:proposed}

\begin{figure}[h!]
	\centering
	\includegraphics[width=\linewidth]{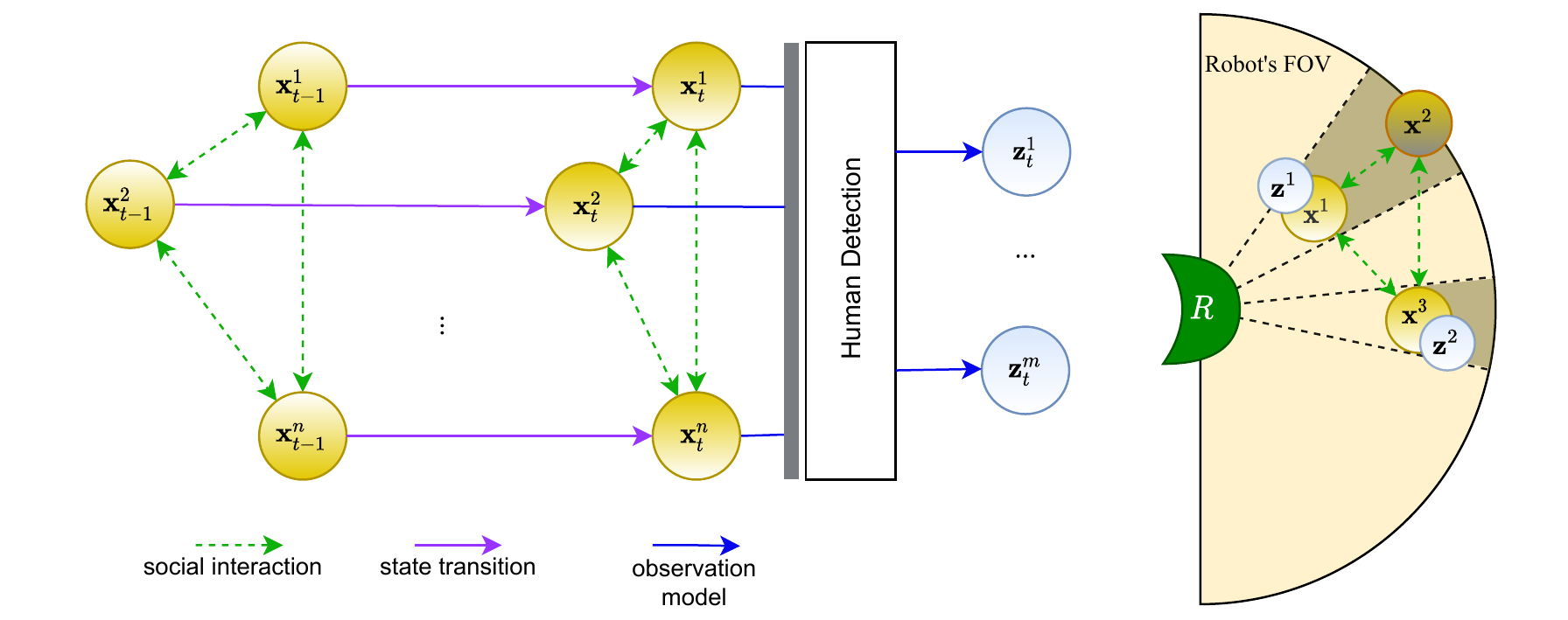}
	\caption{The graphical representation for Occlusion- and Socially-aware Object Tracking. The yellow circles show the state of each agent at two consecutive time instants $t-1$ and $t$. The blue circles are the observations at current time instant $t$. The dashed green arrows represent the social interaction between the agents and the blue arrows represent the observation model.}
	\label{fig:multi-social-object-tracking}
\end{figure}

Suppose that a robot navigates in an environment shared by $n$ pedestrians. Each pedestrian state is described through its position $\mathbf{x}_i\in\mathbb{R}^2$ and instantaneous velocity $\mathbf{v}_i\in\mathbb{R}^2$. The robot uses its sensors to get raw measurements and passes them to a human detection unit that returns a set of $m$ detections as ${\mathbf Z} = \{\mathbf z_1, ..., \mathbf z_{m} \}$ (see Fig. \ref{fig:multi-social-object-tracking}), which are handled as noisy observations of $\{\mathbf x_1, ..., \mathbf x_{m} \}$.
We assume there are $u=n-m$ unobserved pedestrians, either due to errors by the detector or because the pedestrians are not visible by the sensors, e.g. because of an occlusion or because they are out of the robot field of view.

Our algorithm leverages the information about geometric relations between the people in the surrounding crowd, extracted from previously observed trajectories, to infer a probabilistic occupancy map covering occluded regions and to impute the position of unobserved pedestrians. This way, we can improve the robot knowledge about the environment and build more reliable motion plans estimate and plan beyond what it can see. Before detailing our imputation algorithm, we first introduce the concept of Social Tie.

\subsection{Social Ties}
\label{subsec:ST}

As a modeling tool to understand the geometrical structure of the flows within crowds, we introduce the notion of Social Tie.   Inspired by ideas from other works~\cite{alahi2014SAM}, we define a Social Tie as a displacement vector between a pair of agents, expressed in the local frame of the first one: 

\begin{equation}
	\delta({\textbf{x}}_i, {\textbf{x}}_j) = (\textbf{R}_{i})^\top ({\textbf{x}}_j - {\textbf{x}}_i),
	\label{eq:social-tie}
\end{equation}

\noindent
where $\textbf{R}^t_{i}$ is the rotation matrix giving the global orientation angle of agent $i$ at time $t$.

We further classify the social ties as \emph{strong ties} and \emph{absent ties}. The terms are borrowed from social networks literature~\cite{granovetter1973ties} to represent, respectively, long-term interactions (e.g. grouping or leader-follower behaviors) versus instant interactions (e.g. collision avoidance overtaking) between pair of agents.
The motivation behind this modeling choice is that these two categories (strong/absent) define two distributions of displacements $\delta$, that are better treated dissimilarly. 

\begin{figure}[!h]
	\centering
	\includegraphics[trim= 6 25 6 25, clip, width=0.9\columnwidth]{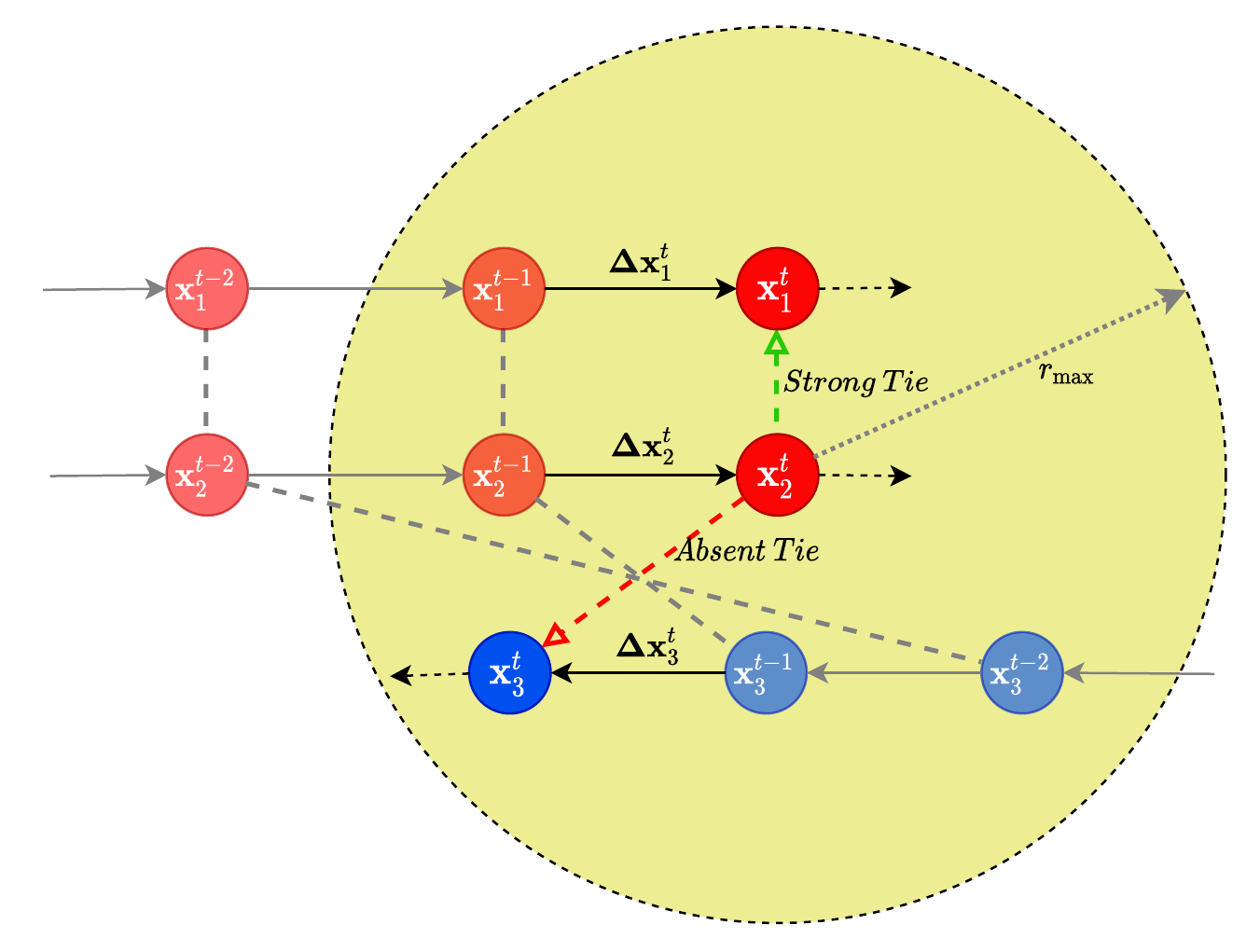}
	\caption{
		Social Ties. The tie between $\mathbf{x}_1$ and $\mathbf{x}_2$ is classified as a \textit{strong} tie at time $t$, (shown in green). On the other hand, the link between $\mathbf{x}_2$ and $\mathbf{x}_3$ is \textit{absent} (red dashed line)
%		 due to significant change from $t-1$ to $t$
		.
	}
	\label{fig:ties_classification_rule}
\end{figure}

The classification between the two tie types is based on the history of the distance between two agents. We keep the history of the distance vector between all pairs of detected agents. 
A tie is labeled as \emph{strong}, if 1) during the last $t_{c}$ time-steps, the two agents are closer than a threshold distance $r_{max}$, 2) in the same time interval, the Euclidean distance between the agents remains fixed (up to a tolerance threshold $\epsilon_l$ on distance variations) over the last second, and 3) the agents have similar orientations (within another threshold $\epsilon_{\theta}$). If 1) holds but either 2) or 3) does not, then the link is categorized as absent.
While we found these simple rules to be enough in our setup, more advanced classification rules can also be used. An example is depicted in Fig.~\ref{fig:ties_classification_rule} to illustrate the classification process.

The above definition implies that a tie (strong or absent) is assigned to any pair of agents that have a distance lower than $r_{max}$ during the last $t_{c}$ time-steps.
It is worth observing that the tie definition is symmetric: a strong (absent) tie from $\textbf{x}_i$ to $\textbf{x}_j$ implies a strong (absent) tie from $\textbf{x}_j$ to $\textbf{x}_i$. However this symmetry property is not directly implied by the definition of social tie, but by the classification rule, which means with a new classification method, one-way ties may exist.

%\CommentJvd{In the last 1.0 sec, no change in distance more than 0.5 meter, and with difference between orientations not more than 45 degree}

Next, we evaluate the distribution of social ties across agents by classifying and accumulating the observed ties and then taking a polar histogram for each. The two histograms represent the empirical distribution of the strong and absent ties. They are denoted by $p(\delta|\tau=S,\mathcal{H})$ and $p(\delta|\tau=A,\mathcal{H})$, respectively, where $\tau$ is the tie type (strong or absent) and $\mathcal{H}$ denotes the historical observations used for training.
Note that these distributions on displacements give us access to the conditional distribution of the absolute location of a queried position $\textbf{x}$, conditioned on seeing a pedestrian at $\textbf{x}_i$, and on the type of social tie $\tau$: $p(\textbf{x}|\textbf{x}_i, \tau, \mathcal{H})$.

\begin{figure*}
	\begin{center}
		\begin{tabular}{cc}
			\includegraphics[trim=1pt 0 1pt 0, clip, width=0.48\textwidth]{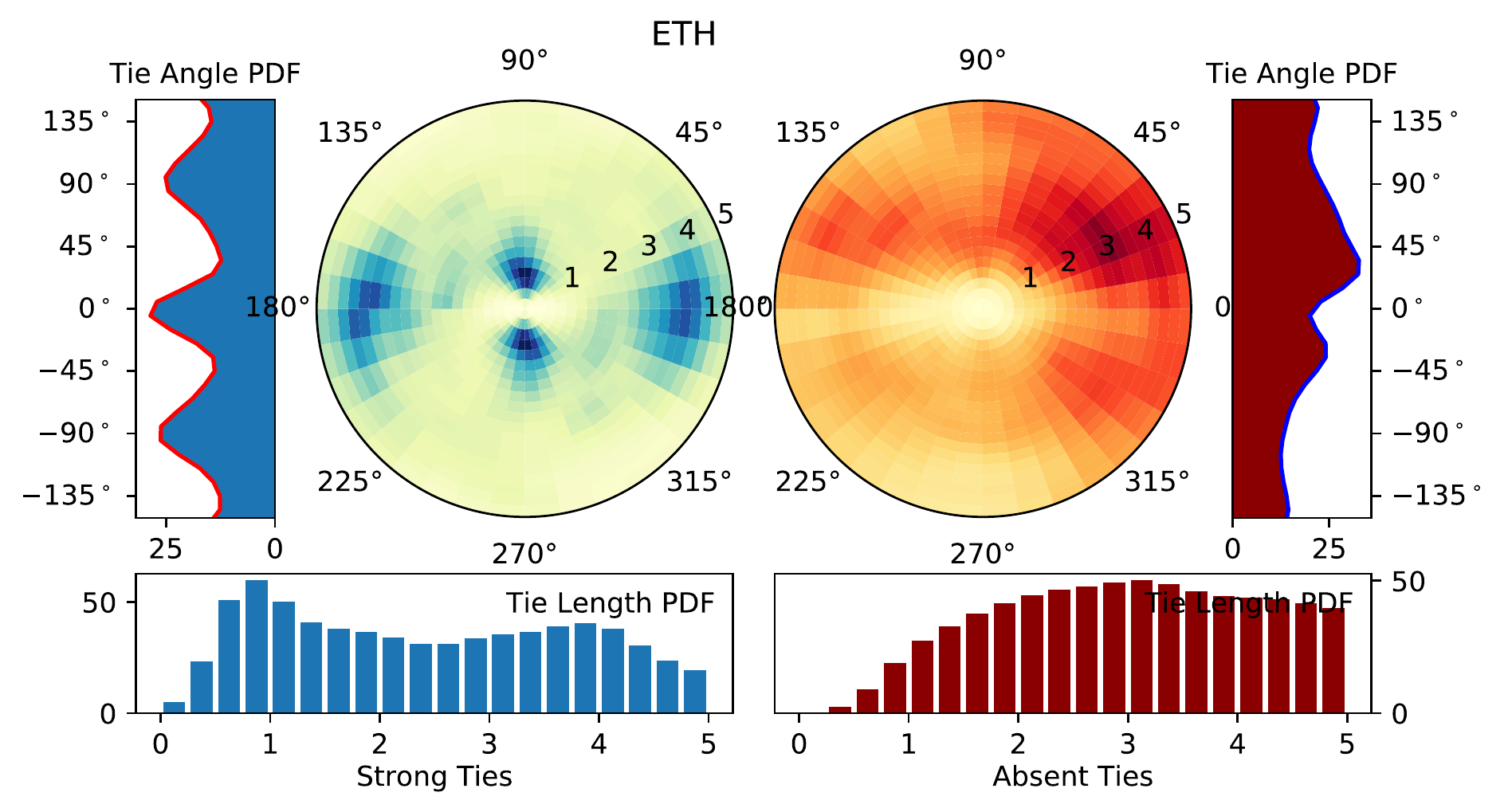} 
			\includegraphics[trim=1pt 0 1pt 0, clip, width=0.48\textwidth]{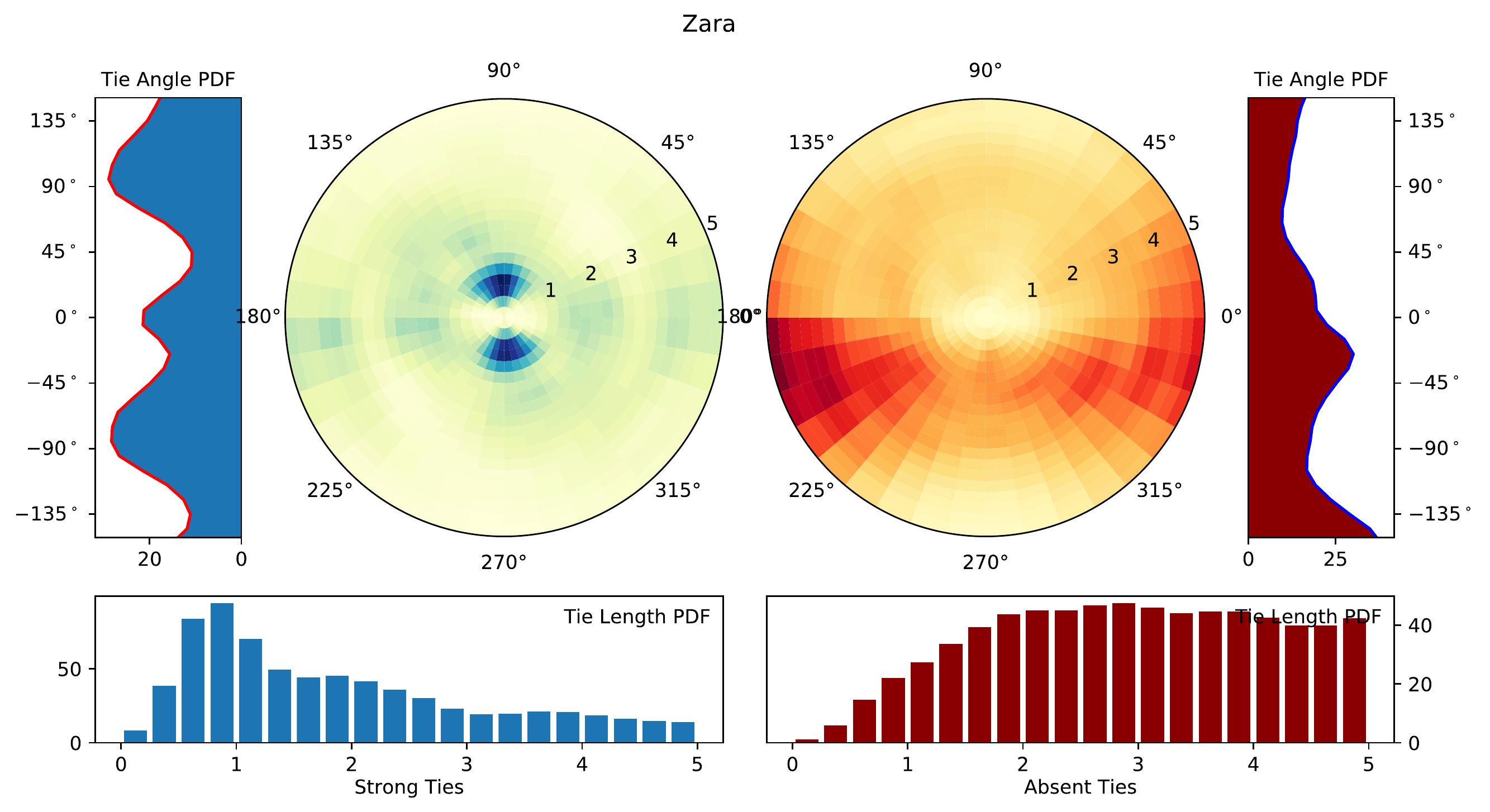} \\
			\includegraphics[trim=1pt 0 1pt 0, clip, width=0.48\textwidth]{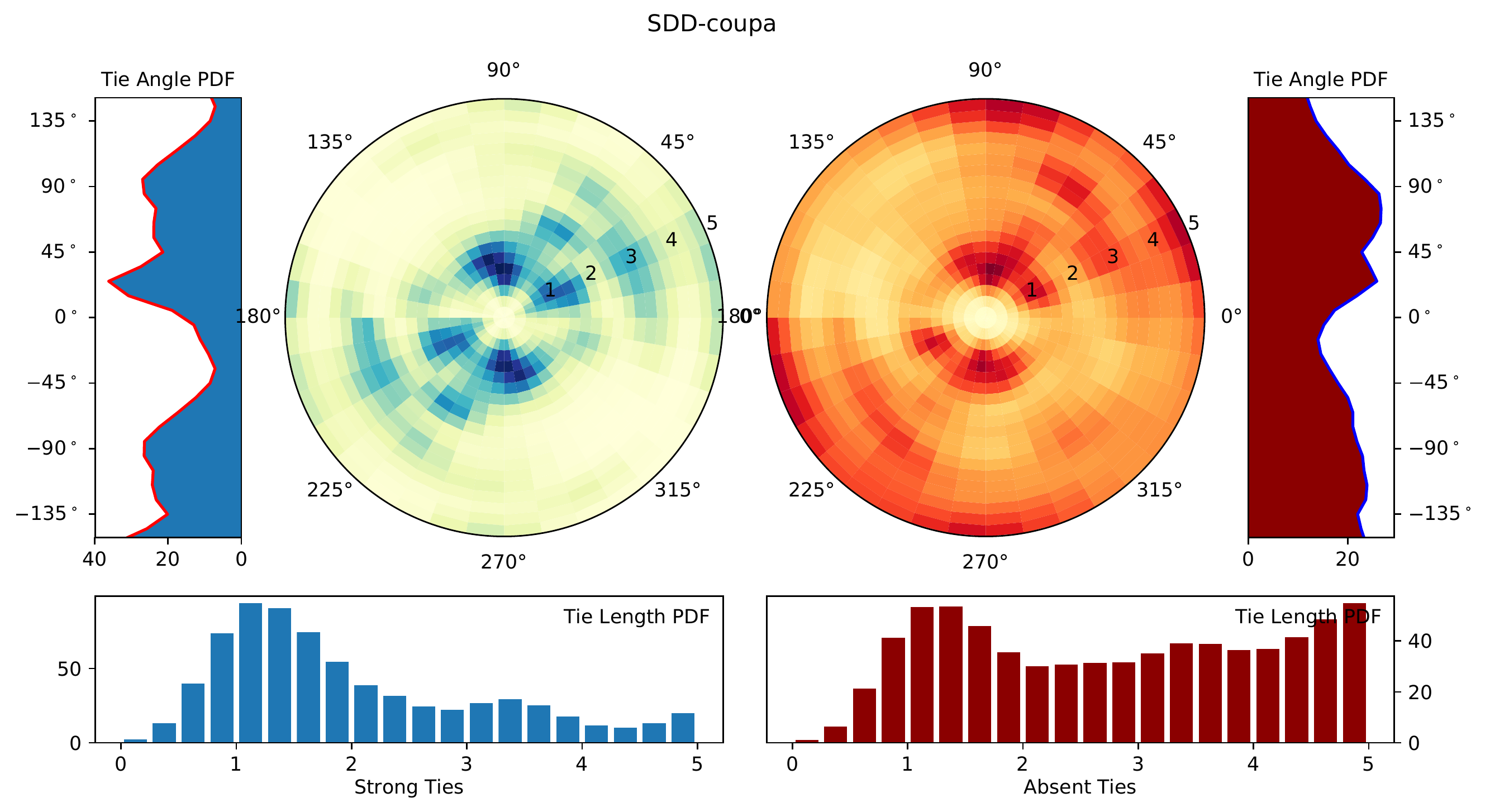} 
			\includegraphics[trim=1pt 0 1pt 0, clip, width=0.48\textwidth]{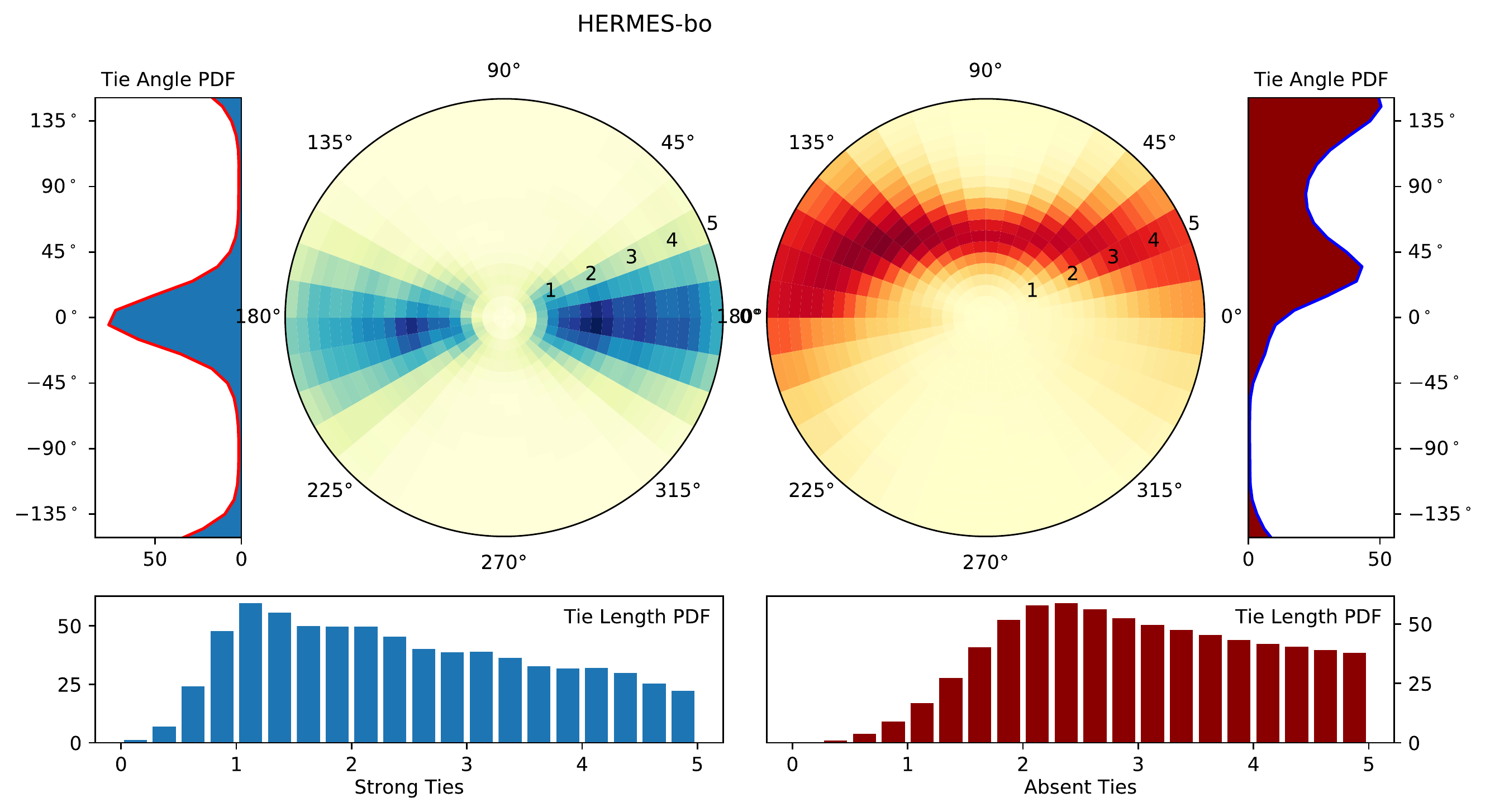}
		\end{tabular}
	\end{center}
	\caption{Distributions $p(\delta|\tau=S,\mathcal{H})$ and $p(\delta|\tau=A,\mathcal{H})$ of \textit{strong} (left side) and \textit{absent} (right side) social ties for 4 different datasets. The different flow structures lead to very distinct distributions along the datasets. The graphs emphasize some structural elements of the crowd flow, such as the communities width (very small in the Hermes-Bottleneck case), the permanent asymmetries of the absent ties (Zara or Hermes-Bottleneck).\label{fig:p-ties}}
	
\end{figure*}

Some examples of $p(\delta|\tau,\mathcal{H})$ are shown in Fig.~\ref{fig:p-ties}, where one can observe that the different collective patterns lead to significantly different social tie patterns in the HERMES-Bottleneck, SDD, ETH and Zara Datasets. For example, in Zara (upper right), the strong ties are mostly observed as side-by-side configurations, while in HERMES-Bottleneck (bottom right), the strong ties come mostly as lines of people (because of the nature of the dataset). We will discuss in the following how these patterns can help us in estimating the location of people in unseen (occluded) areas.

\subsection{Communities (clusters)}
\label{subsec:Flows}

\begin{figure*}
	\begin{center}
		\begin{tabular}{cc}
			\includegraphics[trim= 110 0 90 50, clip, width=0.535\linewidth]{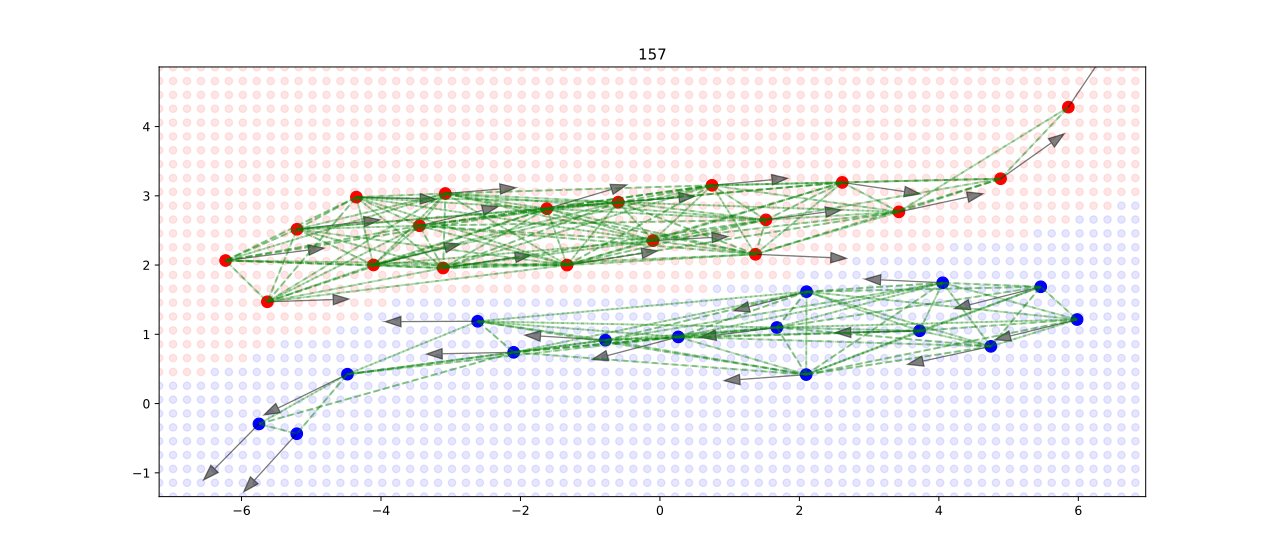}

			\includegraphics[trim= 60 0 60 95, clip,  width=0.44\linewidth]{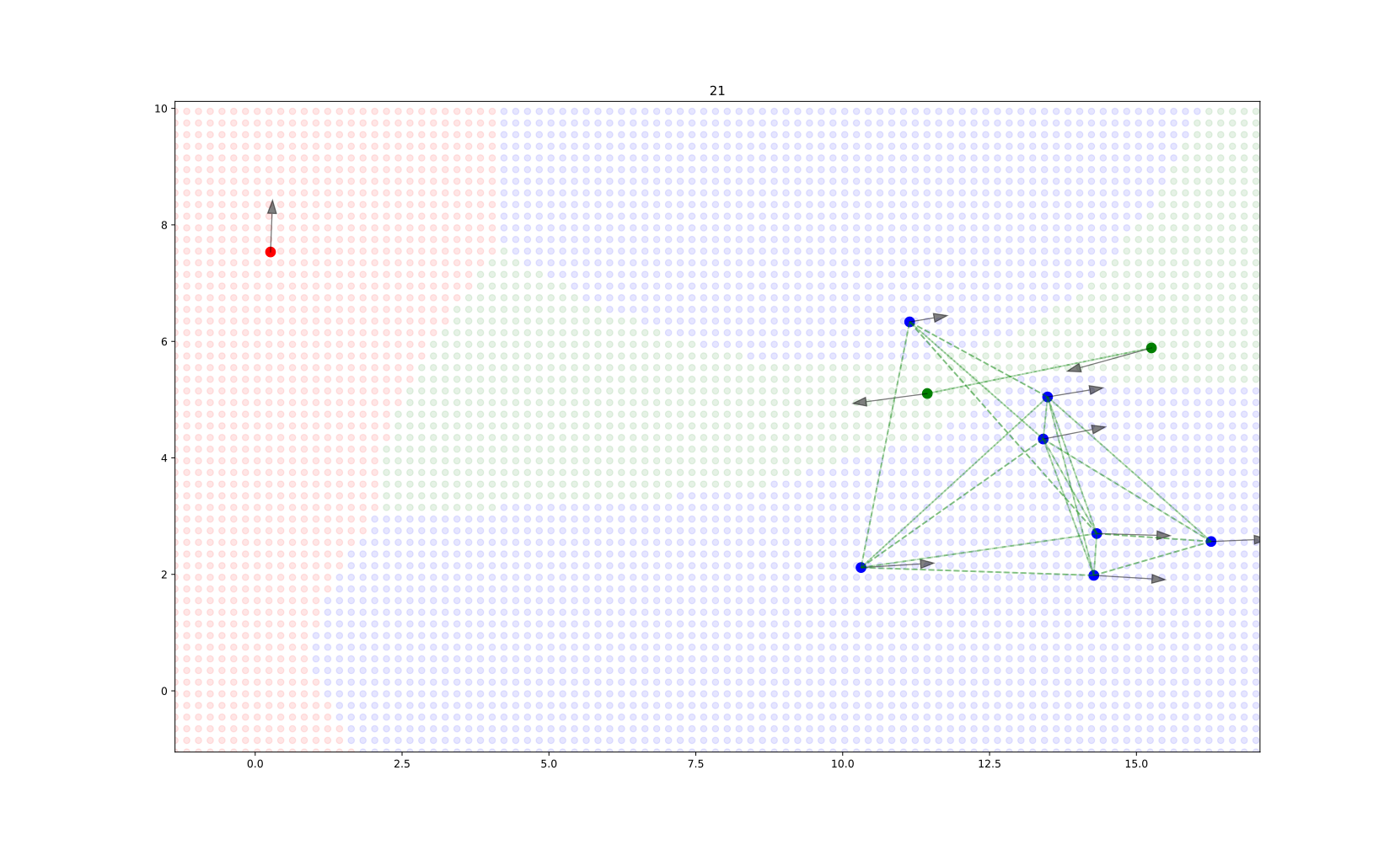}

		\end{tabular}
	\end{center}
	\caption{Communities. Green lines: strong ties, black arrows: velocity vector of the agents. Left: Bottleneck dataset, a bi-directional flow of pedestrians that form two communities, Right: A frame of UCY dataset (Zara) with 3 communities moving in different directions. The territory of each community is shown with different colors on a mesh.}
	\label{fig:communities}
\end{figure*}

We define \textit{communities} (or groups) as subsets of pedestrians that are connected (directly or indirectly) by strong ties, i.e., that can be seen as clusters of people moving together as a group, or as a continuous flow of people moving in one direction (see examples in Fig.~\ref{fig:communities}).
The $m$ observed agents are partitioned into $K$ communities: $\textbf{C} = \{\textbf{c}_1, ..., \textbf{c}_K\}$. Note that a community can be as small as containing only one person.
The velocity of a community $\textbf{c}_k$ is defined as the average velocity of its members:
\begin{equation}
	\bar{\textbf{v}}_k = \frac{1}{|\mathbf c_k|} \sum_{i \in \mathbf c_k}{\textbf{v}_i}
\end{equation}

We also define a \textit{territory} area for each community, by running a k-nearest neighbor classifier  that assigns, to every location in the plane, a label that represents the nearest community. The probability of being, at a location $\mathbf x$, in the territory of $\mathbf c_k$ is denoted by $\phi_{k}(\mathbf x)$. To take the orientation of each agent into account, the k-nearest neighbor is not implemented with the Euclidean distance but with a Mahalanobis distance $d(\mathbf x,\mathbf y)$ with a $2\times 2$ covariance matrix $\Sigma$ chosen with its first eigenvector aligned with the agent velocity ${\textbf{v}}_i$ and assigned to an eigenvalue $\alpha \|{\textbf{v}}_i\|+\beta$ while the second one is assigned an eigenvalue $\beta$:

\begin{align}
d(\mathbf x,\mathbf y)= (\mathbf x-\mathbf y)^T \Sigma^{-1} (\mathbf x-\mathbf y).\label{eq:mahalanobis}
\end{align}

%By classifying social ties between agents as we have shown above, we construct multiple connected components that move in a \textit{crowd flow}. 
%\CommentJB{Only strong ties?}.
%\CommentJvd{Yes. A set of agents that are connected with strong ties (Fig. \ref{fig:classify-ties})}
%The set of connected agents, assigned to $\Phi_i$ 
%\CommentJB{What is phii? }is shown by $\{X^{\Phi_i}\}$ \CommentJB{I dont understand the notation}. We then use a nearest-neighbor classifier to assign the flow class of the nearest agent to a given location $\textbf{x}$. This categorical variable which is shown by $\Phi(\textbf{x})$ will help us to predict and generate virtual agents, explained the next section.
%
%\CommentJB{Some confusion about the notion of classification: Above you described a classification into strong/absent. Is it the same you are talking about here?}

%\begin{figure}
%\centering
%\includegraphics[width=0.9\linewidth]{figs/flow/crowd_flow_map}
%\caption{Above: ground truth agents from Bottleneck Dataset. Bottom: detected and tracked agents by the robot (black circle) and two classes of flows shown by red and blue.}
%\label{fig:crowd_flows}
%\end{figure}

\subsection{Imputing New Pedestrians}
\label{subsec:Virtual}

We have explained in~\ref{subsec:ST} that the social ties are extracted from crowd activity observations. Here we leverage this data to predict plausible positions for the unobserved agents that might exist in the blind spot areas and beyond the field of view of the sensor. This idea is inspired by \textit{inpainting} technique in Computer Graphics, where Pair Correlation Functions (PCF)~\cite{nicolet2020pcf} are used to detect textures from an image and propagate them to other areas.

A PCF measures the probability density of the distance between pairs of particles. However, in our context, considering only the distance between agents is difficult, since the orientation of the agents is critical in modeling human collective activities~\cite{moussaid2010walking}. Hence, we propose an extension of the concept of PCF with the distribution of social ties.

We are interested in finding 
$p(o(\mathbf{x})|\mathbf{Z})$, the probability of the presence of an unobserved agent 
at a query point $\textbf{x}\in \mathbb{R}^2$, given the $m$ detected agents $\mathbf{Z}$.

By expressing this distribution as a marginalization over the two latent variables $v(\mathbf x)$ and $\mathbf c(\mathbf x)$ for, respectively, the visibility from the robot at position $\mathbf x$ and the class indicating the community at that position, we can write this as (we removed the $\mathbf x$ for shortening equations, but be aware that the random variables $o$, $c$, $v$ are all defined in a specific $\mathbf x$)

{\small
\begin{align}
	 p(o|\mathbf Z) &=&\sum_{w\in \{0,1\}} \sum_k  p(o , \mathbf  c=\mathbf c_k, v=w |\mathbf{Z})\\
	 &=&p(v=0) \sum_k  p(\mathbf  c=\mathbf c_k) p(o| v=0,\mathbf  c=\mathbf c_k,\mathbf{Z}).
	 \label{eq:marginalized-2}
\end{align}
}

Using the community-dependent ties distributions, we  estimate $p(o(\mathbf x)| v(\mathbf x)=0,\mathbf  c(\mathbf x)=\mathbf c_k,\mathbf{Z})$ as follows:

{\small
\begin{align}
\begin{split}
	p(o| v=0,\mathbf  c=\mathbf c_k,\mathbf{Z}) \propto
	\prod_
	{\mathbf z_i \in \mathbf Z \cap \textbf{c}^k}
	\int_{\textbf{x}_i} p(\delta(\textbf{x}_i, \textbf{x}) |\tau=S,\mathcal{H})p(\textbf{z}_i|\textbf{x}_i)d\textbf{x}_i \\
	\times
	\prod_
	{\mathbf z_j \in \mathbf Z \backslash \textbf{c}^k} 
	\int_{\textbf{x}_j} p(\delta(\textbf{x}_j, \textbf{x}) |\tau=A,\mathcal{H})p(\textbf{z}_j|\textbf{x}_j)d\textbf{x}_j
\end{split}	
\end{align}
}

\noindent where $p(\textbf{z}_i|\textbf{x}_i)$ is the sensor error model and
$p(\delta|\tau,\mathcal{H})$ denotes the likelihood of a social tie $\delta$ (see Section~\ref{subsec:ST}). If the sensor model is Gaussian, the integrals above can be evaluated easily by using pre-filtered versions of the maps $p(\delta(\textbf{x}_i, \textbf{x}) |\tau,\mathcal{H})$. This model factors the joint distribution through pairs of agents.
In practice, we only consider the agents in $\textbf{X}$ within a radius $r_{max}$ from $\textbf{x}_q$.
Rewriting Eq. \eqref{eq:marginalized-2} using the notation introduced above $\phi_{k}=p(\mathbf  c=\mathbf c_k)$:

\begin{equation}
	p(o|\mathbf Z) =
	p(v=0)
	\sum_{k}
	\phi_{k} 
	 p(o| v=0,\mathbf  c=\mathbf c_k,\mathbf{Z}).
\label{eq:product_p(ties)} 	
\end{equation}

By interpreting this distribution as a likelihood function $q(\mathbf x)=p(o(\mathbf x)=1|\mathbf Z)$ over $\mathbf x$ and by discretizing the $\mathbf x$ along a regular grid, we can sample a new agent at location $\mathbf{x}_s$.
We tested different sampling strategies, and found out the following sampling to be more effective: we draw a sample from $q(\mathbf x)$ and then we search within a small disk $r_s$ for the local maximum of $q$ in this region.
After this, a virtual agent is created at this location $\mathbf x_s$, and assigned to community $\argmax_{\textbf{c}_k} p(\textbf{c}(\mathbf x_s)=\textbf{c}_k)$, with velocity $\textbf{v}_k$.
By iterating the sampling, we create more agents in the occluded area. After each sampling iteration $\#i$, we add the sample $\mathbf{x}^{(i)}_s$ to $\mathbf{ Z}^{(i-1)}$ (having $\mathbf{ Z}^{(0)}\equiv\mathbf{Z}$) and update $q^{(i)}$ using Eq. \eqref{eq:product_p(ties)}.
%
%\hl{How many times we do this process?}
We repeat the process until $\max_{\textbf{x}} q^{(i)}(\textbf{x}) < \epsilon_s$ in a neighborhood of radius $r_{nav}$  around the robot.

\subsection{Sampling an Imputed Crowd}
By repeating the sampling process, explained in the previous section, we obtain an augmented set $\mathbf{Z}^+$ of detected and virtual agents. We repeat this entire process, for $H$ times to get multiple sets of hypotheses $\mathbf{ Z}^{+(h)}$ for $h=1...H$.

The pseudo-code for our proposed algorithm "Occlusion-Aware Crowd Imputation" can be found in Algo.\ref{alg:overall}.

\begin{algorithm}
	\caption{Occlusion-Aware Crowd Imputation}
	\label{alg:crowd-pred-init}
	
	\hspace*{\algorithmicindent} 
	\textbf{Inputs}: ${\mathbf Z} = \{\mathbf z_1, ..., \mathbf z_{m} \}$
	\Comment{detected persons} \\
	%\hspace*{\algorithmicindent} \hspace*{31pt} $\textbf{\^X}^t$
	%\Comment{predicted tracks from $t-1$} \\	
	\hspace*{\algorithmicindent} 
	\textbf{Output}: Sample sets $\{\mathbf{Z}^{+(h)}\}_h$
	
	\begin{algorithmic}[1]
		%\State Assign detections $\mathbf Z$ to tracks $\textbf{\^X}^t$ using \textit{Munkres} algo.
		
		%\State Initialize new tracks for unassigned detections
		
%		\State \hl{Change} status of Virtual tracks to Normal if they got assigned
		
	%	\State %$\textbf{X}^t \leftarrow$ 
		%Update tracks based on assigned detections 
%		\CommentJB{Dont understand the notation with the arrow}
%		\CommentJvd{Assignment :-)}
		
		\State Compute Ties and Tie types between detections (Eq. \ref{eq:social-tie})
%		\State Update $p(\delta^S|\textbf{Z}^{1:t})$ and $p(\delta^A|\textbf{Z}^{1:t})$ Eq. (Todo: explain prior distribution and update policy)
		
		\State Find Connected Components (Communities) $\{\textbf{c}_1 .. \textbf{c}_K\}$
		
		\State Compute Territory of each Community $\{\phi_{1} .. \phi_{K}\}$

		\State Compute Velocity of each Community $\{\bar{\textbf{v}}_1  .. \bar{\textbf{v}}_K\}$
		
		\State Initialize: $\forall \mathbf x$  $q(\textbf{x}) \leftarrow p(v(\textbf{x})=0)$
		
		\For {$\mathbf x$}
			\State $k'\leftarrow \mathbf c(\mathbf x)$ 			
			\For {$\textbf{z}_i \in \mathbf Z$}	
				\State $\delta \leftarrow  \textbf{R}_i^T(\mathbf x-\mathbf z_i)$
				\State $k\leftarrow \mathbf c(\mathbf z_i)$ 
				\If{$k=k'$}
					\State $q(o) \leftarrow q(o) * p(\delta|\tau=S,\mathcal H)$
				\Else 
				 	\State $q(o) \leftarrow q(o) *  p(\delta|\tau=A´,\mathcal H)$ 
				\EndIf
			\EndFor
		\EndFor

		\For {$h \in {1..H}$}
			\State $\mathbf{Z}^{+(h)} \leftarrow \mathbf Z$
			\While {$\max q(\mathbf x) > \epsilon_s$}
				\State $\textbf{x}_s \leftarrow$ Sample virtual pedestrian 
				\State add $\textbf{x}_s$ to $\mathbf{Z}^{+(h)}$
				\State assign $\textbf{x}_s$ to $\textbf{c}_k=\argmax_{\textbf{c}} p(\textbf{c}(\mathbf x_s)=\textbf{c})$
				
				\State update $q(\textbf{x})$ for all $\textbf{x}$ (Lines 6:17)
				
			\EndWhile
		
		\EndFor

%		\State \hl{Refine} virtual agents $\{\textbf{x}_v\}$ using Eq. \ref{eq:product_p(ties)}
		
%		\State Predict jointly the state (trajectory) of all agents (real+virtual) for $t+1:t+T$ Eq. \ref{eq:social-predict}  for each hypothesis
		
		\State Return $\{\mathbf{Z}^{+(h)}\}_h$
		
		%		\For {$t \in [1, \infty)$}	
		%		\EndFor
	\end{algorithmic}\label{alg:overall}
\end{algorithm}

%	\item Confirm tracks if they have appeared in $\Gamma\%$ of recent frames
%	\item Delete tracks if they have remained unassigned for too long, while being posed within FOV of the robot. \hl{Explain here more clearly}
%\end{itemize}

%%\input{pcf_gp}
%%\input{followbot}

% !TeX root=main.tex

\section{Experimental Results}
\label{sec:experiments}

In this section we first study the feasibility/importance of the proposed method on multiple trajectory datasets and then we discuss the results.
%We evaluated the proposed method on both simulated and real data.
%\subsection{Where this solution can be applied?}

%1. check performance in different Crowd local density (centered on the Robot)
%2. the variance of the flows (test in Unidirectional scenario / Bidirectional and multiple-directional flows)

\subsection{Implementation Details}
We developed our algorithm using Python. For 3D simulation of the crowd motions and the robot perception, we used the CrowdBot Simulator, which is built on top of Unity game engine and ROS system. Perceptions are obtained from a simulated 360-degree LiDAR with a resolution of $0.5^o$ and working range of $[0.05m, 8m]$ installed at a height of $40cm$. We use DR-SPAAM~\cite{Jia2020DRSPAAM}, a deep learning-based person detector that detects persons (legs) in 2D range data sequences. We couple the detector to a Const-Acceleration Kalman-Filter to track multiple targets. In our algorithm, the gridmap has a resolution of $8$ cells per meter.

The hyper-parameters of the algorithm are chosen as follows: $r_{max} = 5m$, $t_c = 1s$, $\epsilon_l = 0.5m$ and $\epsilon_\theta = 45^o$ for classifying the ties. The polar histogram used for the representation of strong and absent tie distributions have radial and angular resolution of $25cm$ ($[0-5m]$) and $10^o$ ($[-180, 180^o]$) respectively, with a constant-padding $p=1$ for any bin out of this range. Also, the coefficients $\alpha$ and $\beta$ used in Eq. \eqref{eq:mahalanobis} are set to $0.2$ and $0.1$ respectively. $\epsilon_s$ is set to $50\%$ and $r_s$ (the radius of the search disk) is set to $2m$.

%\textit{Implementation tips:}
%The distributions are calculated using polar histograms over the observed ties. We only keep the ties at a distance shorter than a radius $R_{max}$. An example distribution is shown in Figure \ref{fig:p_link} for a simulated crowd that is composed of bi-directional flow of agents, moving in groups of size 2.

%The product in the equation above is performed after rotating the polar histogram $p(\delta)$ toward direction of each agent and then converted to Cartesian form. In the areas out of histogram bound, we pad the probability with $\bar p = mean(p(\delta))$. \CommentJB{Not very clear.}
%In Alg.~\ref{alg:crowd-pred-init}, we describe the overall process of crowd synthesis and prediction for the robot. 
%The inputs of this process are two Probabilistic Occupancy Maps (POM) $\pi_0$ (from the data) and $\pi^t_{r}$ (from the robot's Lidar at time $t$).

\subsection{Real Crowd + Simulated Robot}
To evaluate the proposed method and validate the system, it is critical to work with real datasets. However, to the best of our knowledge, there is no public robot-crowd datasets in which the ground truth annotations of occluded pedestrians are available. Hence, we use crowd-only datasets, select one random pedestrian in the crowd and replace it by a simulated robot. The robot traverses the same trajectory taken by the pedestrian. The motivation for not simulating the robot navigation is to make the system independent of the navigation algorithm and to be able to compare the performance against new methods.

%\begin{figure}
%	\centering
%	\includegraphics[width=0.9\linewidth]{figs/real-test-Hermes}
%	\caption{Hermes Dataset: The robot which replaces a real pedestrian is shown by black circle. Other agents are shown by blue.}
%	\label{fig:real-test-Hermes}
%\end{figure}

\subsection{Crowd Datasets} 
We consider multiple Human Trajectory datasets that cover different crowd densities and crowd structures:

\begin{itemize}
	\item \textbf{SDD} or Stanford Drone Dataset contains trajectories of moving agents in the campus of Stanford University, mostly with low crowd densities \cite{SocialEtiquette2016}.
	
	\item \textbf{ETH} is a small dataset of pedestrians entering/exiting the entrance of a university building \cite{Pellegrini2009}.
	
	\item \textbf{Zara} is a dataset of crowd activity in sidewalk of a shopping street (subset of UCY \cite{CrowdsByExample2007}). 
		
	\item \textbf{Hermes} includes multiple high-density crowd controlled experiments (e.g., through Bottlenecks) in Unidirectional or Bidirectional settings \cite{bottleneck-seyfried}.
\end{itemize}

We use the OpenTraj toolkit~\cite{amirian2020opentraj} to load and process HTP datasets. We split each dataset, into a training set (for extracting social patterns and tie distributions $p(\delta)$) and a testing set, using the 70:30 rule of thumb.

\subsection{Occlusion Severity in Crowd}
\begin{figure*}
	\begin{center}
		\begin{tabular}{cccc}
			\includegraphics[trim=20pt 20 40pt 20, clip, width=0.23\textwidth]{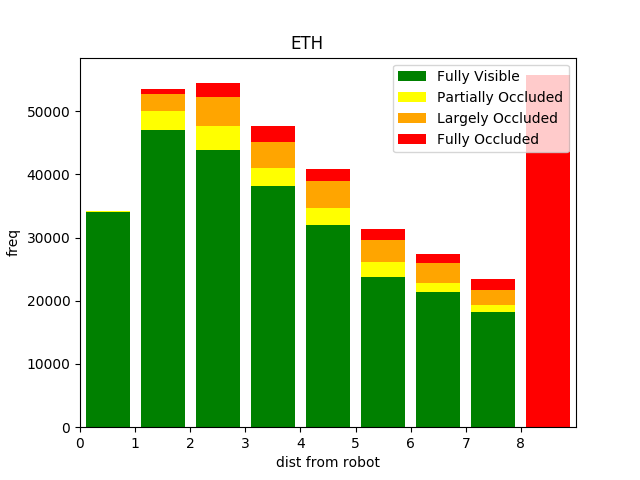}& 
			\includegraphics[trim=20pt 20 40pt 20, clip, width=0.23\textwidth]{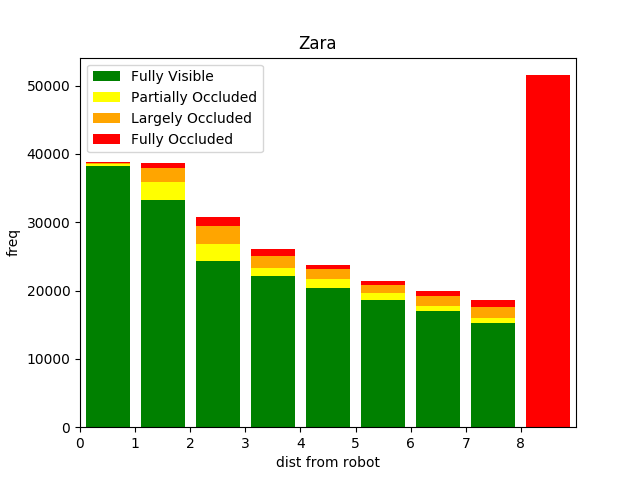}&
			\includegraphics[trim=20pt 20 40pt 20, clip, width=0.23\textwidth]{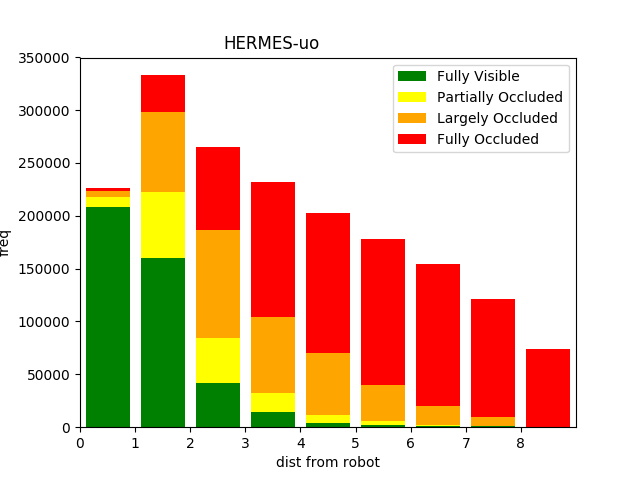}&
			\includegraphics[trim=20pt 20 40pt 20, clip, width=0.23\textwidth]{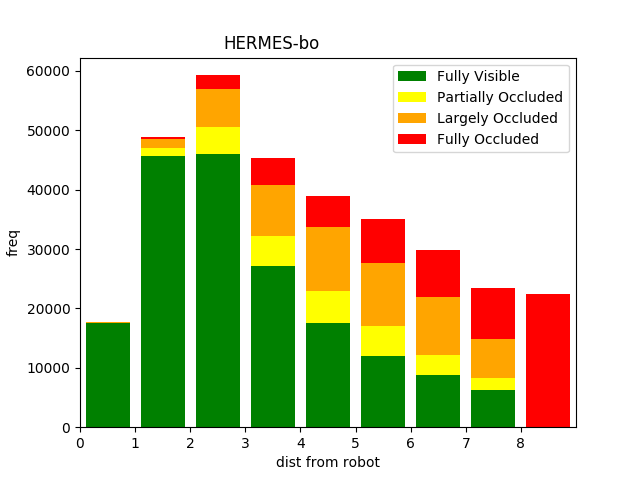}
		\end{tabular}
	\end{center}
	\caption{Occlusion Severity: ETH / Zara / Hermes (Uni-directional and Bi-directional flows)}
	\label{fig:occlusion-severity}
\end{figure*}

We have measure the severity of occlusions for multiple datasets, using the simulation explained above. This is done by repeating the simulation, each time replacing one pedestrian with the robot. Then we estimate the percentage of sensing rays that are occluded for each pedestrian. We have classified occlusion values into to 4 categories: Fully-Visible (0-15\%), Partially-Occluded (15-50\%), Largely-Occluded (50-85\%) and Fully-Occluded (85-100\%). The results are shown in Fig.~\ref{fig:occlusion-severity}. As expected, the HERMES sequences exhibit severe occlusions, while ETH/Zara do not. We will see in the next sections that our crowd imputation algorithm does not improve the baseline in low/medium-density situations.

\subsection{Analysis of Tie Patterns}
\begin{figure}
\centering
\includegraphics[trim=0 30 0 60 ,clip, width=0.9\linewidth]{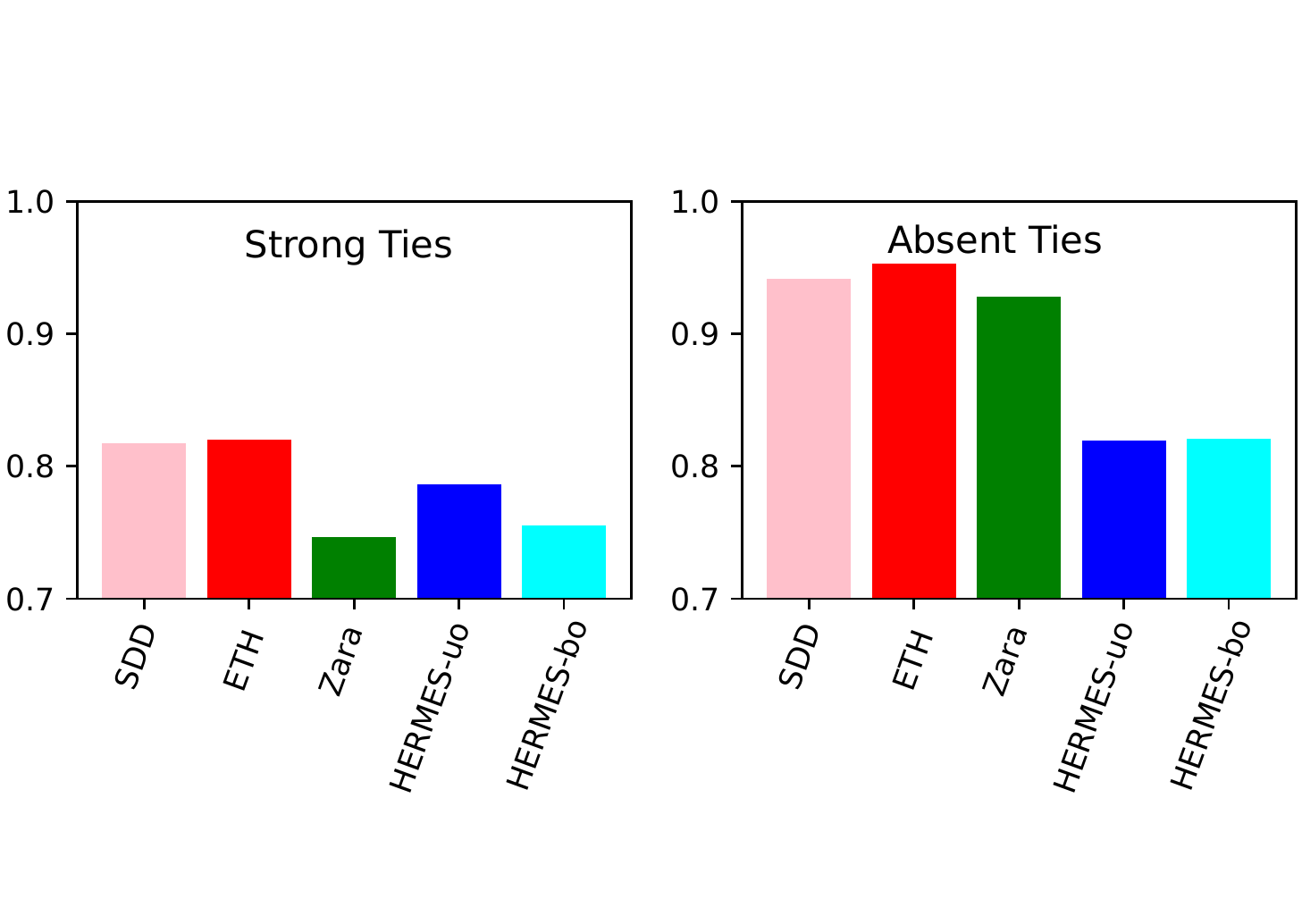}
\caption{Entropies of Strong/Absent Ties distributions for different datasets. Lower entropy means the distribution contains more structured patterns.}
\label{fig:Entropies}
\end{figure}

In order to measure the amount of information captured by the tie patterns, we calculate the entropy of each pattern for each dataset. The equation of normalized entropy for the distribution is derived by considering different bin sizes in a polar histogram~\cite{wallis2006note}:

\begin{equation}
\bar H(p) = -\sum_{r, \theta} 
		p^{r, \theta}
        \log \left( \frac{p^{r, \theta}}{A^{r, \theta}} \right) / H_{max}        
\end{equation}

\noindent
where $p^{r, \theta}$ and $A^{r, \theta}$ are the normalized value and the area of the bin $(r, \theta)$. The total value is divided by $H_{max} = \log (A^{R, 2\pi})$, the maximum entropy of a uniform disk to obtain a normalized entropy between $[0, 1]$. 
In Fig. \ref{fig:Entropies}, we see the entropy values for different datasets.

%\begin{figure}
%\centering
%\includegraphics[width=0.7\linewidth]{figs/links/ETH-Hotel-link-pdf}
%\caption{}
%\label{fig:ETH-Hotel-link-pdf}
%\end{figure}
%\begin{figure}
%\centering
%\includegraphics[width=0.7\linewidth]{figs/links/HERMES-uo-180-180-180-link-pdf}
%\caption{}
%\label{fig:HERMES-uo-180-180-180-link-pdf}
%\end{figure}

%\subsection{Crowd Simulation}
%We simulated a crowd scenario in a narrow corridor using RVO algorithm equipped with grouping behavior.
%
%We tested on different crowd densities and different group size and configuration. To make the simulation close to real scenarios, we set the parameters of the crowd simulation according to the values reported in \cite{moussaid2010walking}. This data are collected in Tolouse city in France in two low and moderate densities.
%
%\begin{figure}
%	\centering
%	\includegraphics[width=0.9\linewidth]{figs/sim-bidirectional}
%	\caption{Simulation of bidirectional crowd with agents moving in pairs.}
%	\label{fig:sim-bidirectional}
%\end{figure}

\subsection{Baselines}
We consider two baselines to compare with our algorithm:
\begin{itemize}
	\item [1.] \textbf{Vanilla-MOT}: A Multi-object tacking without handling the occluded agents, making no assumption about possible agents in occluded areas.
	\item [2.] \textbf{PCF}: A comparable algorithm for analysis and synthesis of point distributions based on Point Correlation Functions (PCF) \cite{oztireli2012gendartthrowing}. The algorithm uses a target PCF (or distribution), that is extracted by analyzing the historical data $\mathcal{H}$, and tries to reconstruct a given set of points (here: the partially occluded crowd) by iteratively sampling new points and accepting/rejecting them based on matching the source and the target PCF curves.
\end{itemize}

\subsection{Performance Evaluation}
%\subsection{Evaluation Metrics}
In order to evaluate the results, we first perform Kernel Density Estimation from the ground truth distribution of agents location, by using Gaussian kernels centered at the location of each agent with $\sigma=0.5m$. We denote the result (a probabilistic occcupancy map) by $\pi$. The Mean Squared Error (MSE) is calculated by averaging the squared difference of predicted and ground truth occupancy grid maps: 

\begin{equation}
	MSE = 
	\frac{1}{A \times H} % T}
	\sum_{h=1}^{H} 
	\sum_{x, y} 
	%	\sum_{t=1}^{T}
	\left\Vert\pi - \hat\pi^{(h)}\right\Vert^2
\end{equation}

\noindent
where H is the number of generated samples, and A is the area of the map. The map $\hat\pi^{(h)}$ is built similarly to $\pi$, based on the sample $\mathbf Z^{+(h)}$. 

%we define the Binary Cross Entropy (BCE) between the projections and ground truth occupancy map:
%
%\begin{equation}
%	BCE = \frac{-1}{A \times H}
%	\sum_{h=1}^{H} 
%	%	\sum_{t=1}^{T} 
%	\sum_{x, y} 
%	\pi^t({x, y}) \log \hat\pi^t_h({x, y}).
%\end{equation}

\begin{figure*}
	\centering
	\includegraphics[width=0.475\linewidth]{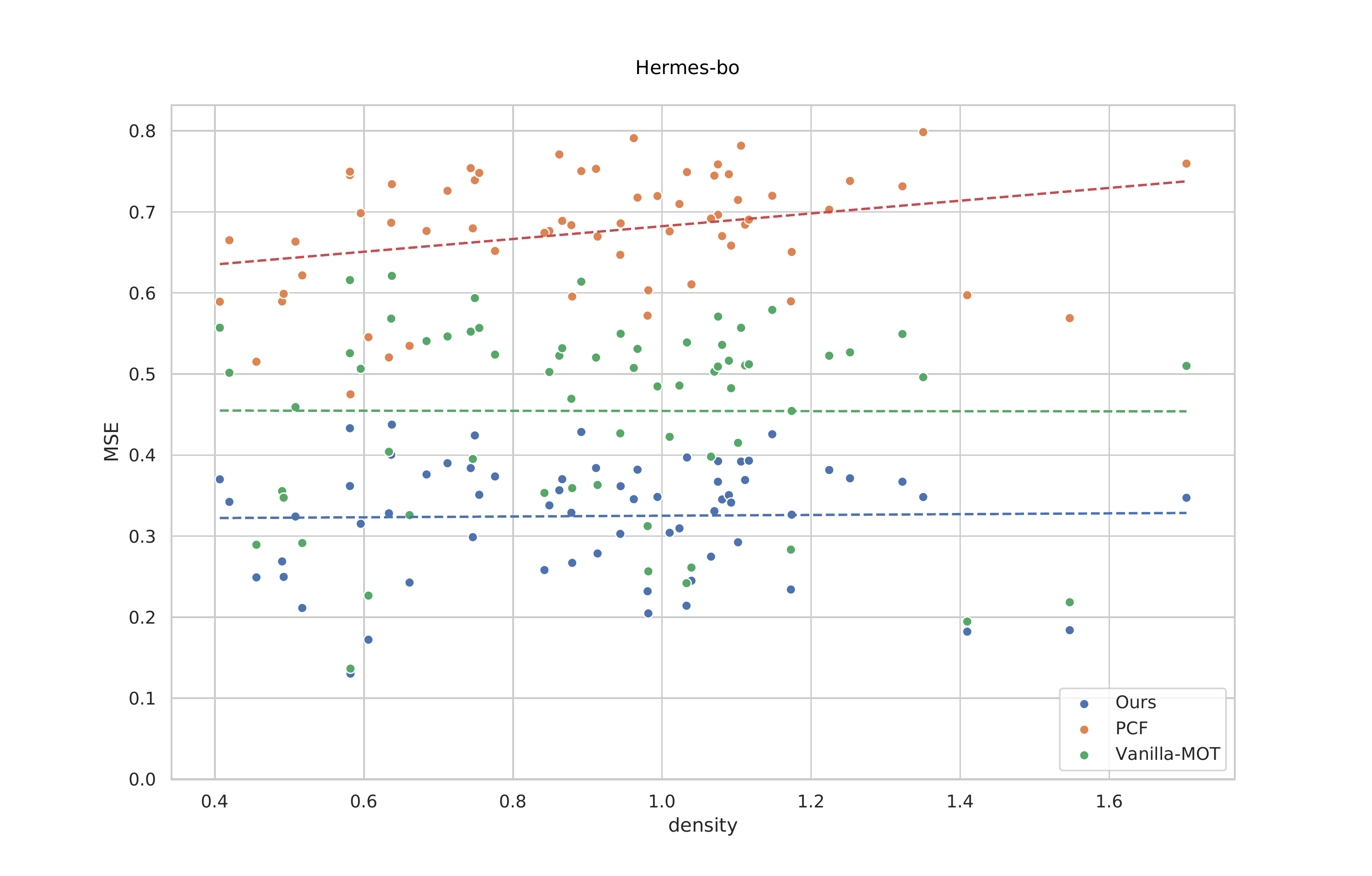}
	\includegraphics[width=0.475\linewidth]{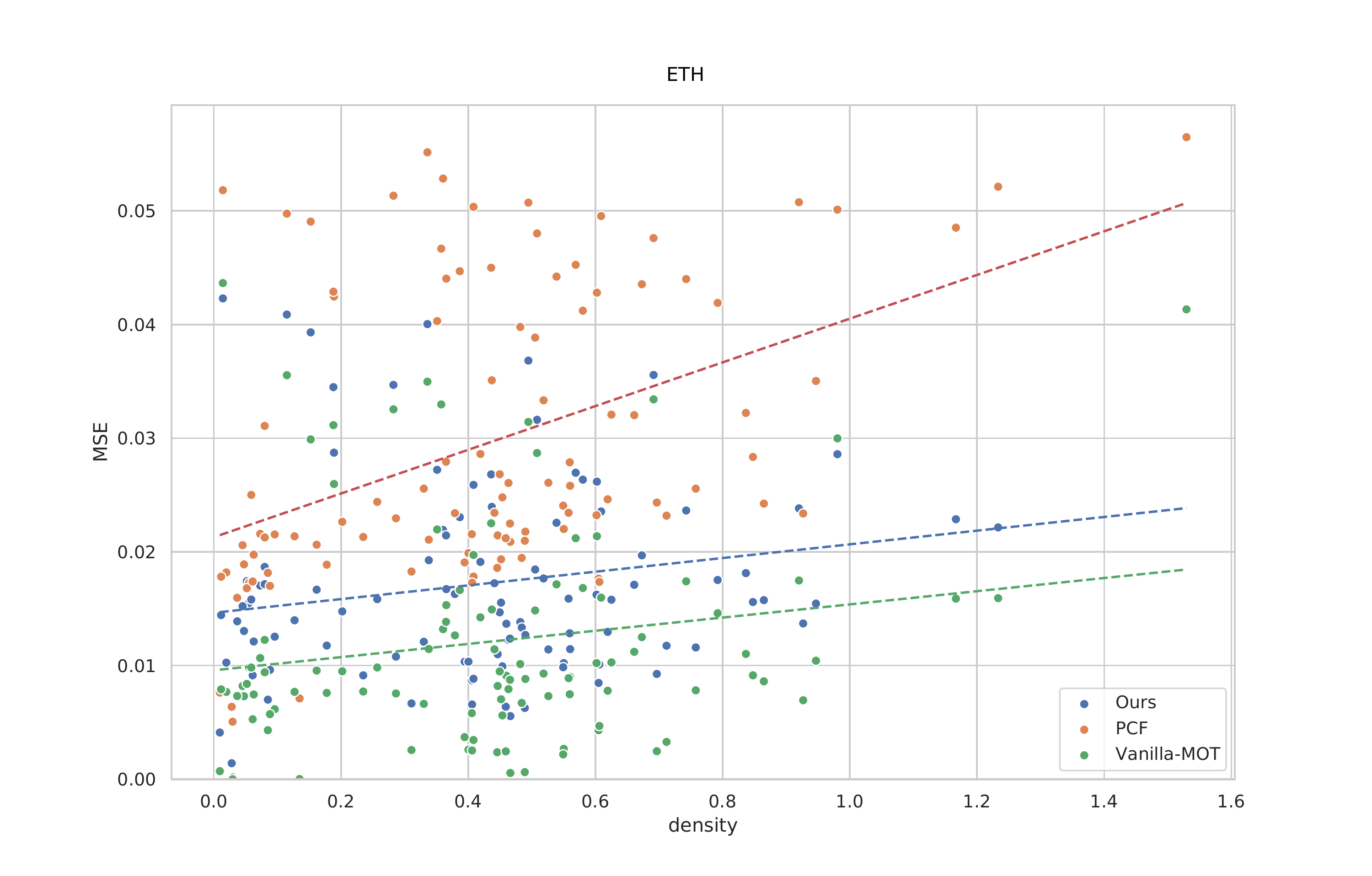}
	\caption{Prediction MSE on Hermes and ETH datasets. Each point on the plot represents the average error of the predictions for one trajectory, sorted by the average crowd density around the robot.}
	\label{fig:Pred-results-MSE}
\end{figure*}

We run the simulation, for each of the trajectory $\#i$ in the test set, and execute the crowd prediction algorithm to obtain $\hat \pi^{(h)}_i$ at each time-step. The prediction errors for Hermes and ETH datasets are shown in Fig. \ref{fig:Pred-results-MSE}. As you can see the proposed algorithm has improved the Vanilla-MOT baseline on Hermes dataset and also outperforms the PCF baseline. On the other hand, on ETH dataset the algorithm has not improved the results which means it's better no make no prediction in scenarios with few-occlusions. Due to this issue we do not report the prediction results on other low-occlusion datasets (SDD and Zara).

\begin{figure*}
	\begin{center}
		\begin{tabular}{cc}
			\includegraphics[trim=80pt 20 80pt 20, clip, width=0.42\textwidth]{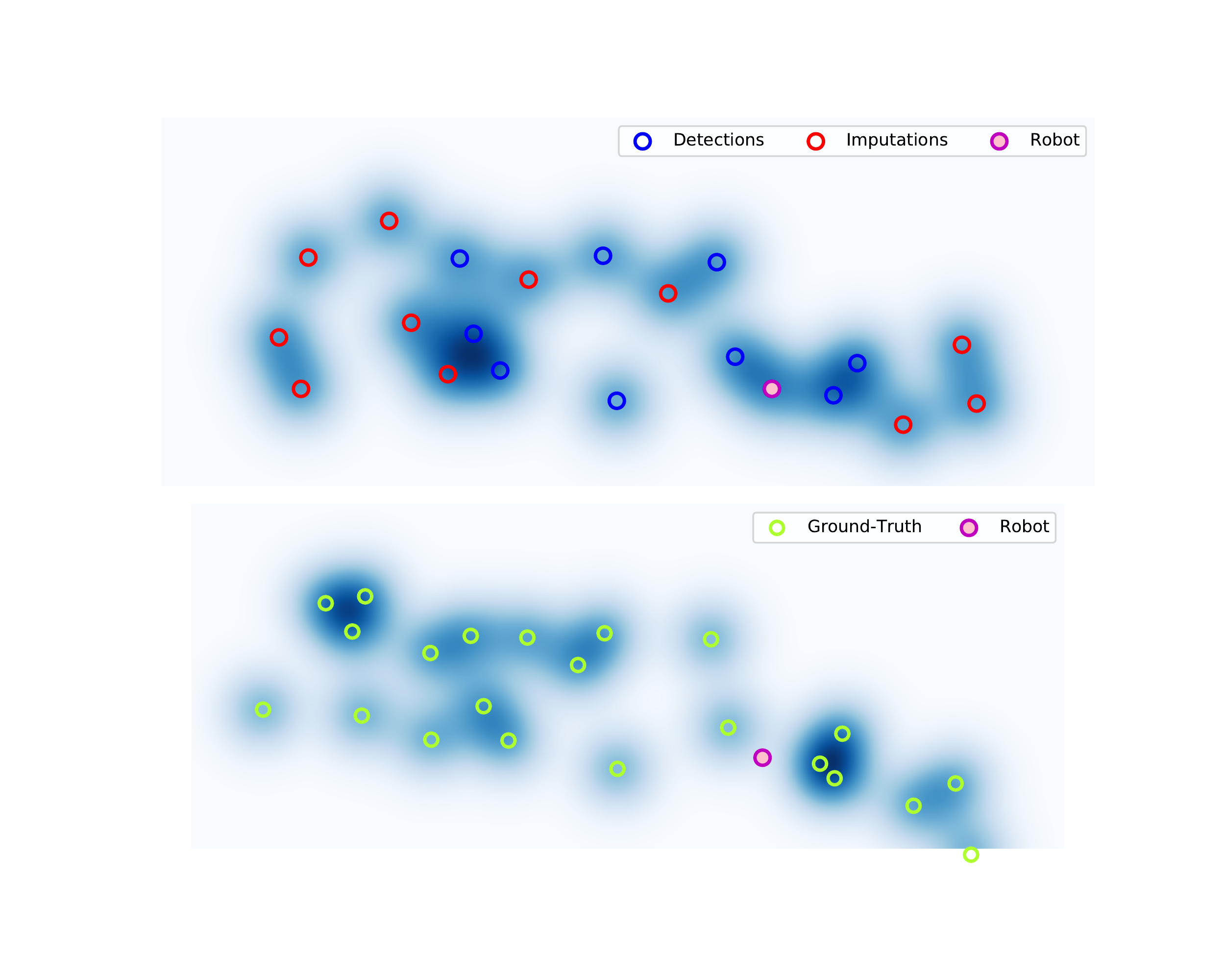}& 
		
			\includegraphics[trim=80pt 20 80pt 20, clip, width=0.42\textwidth]{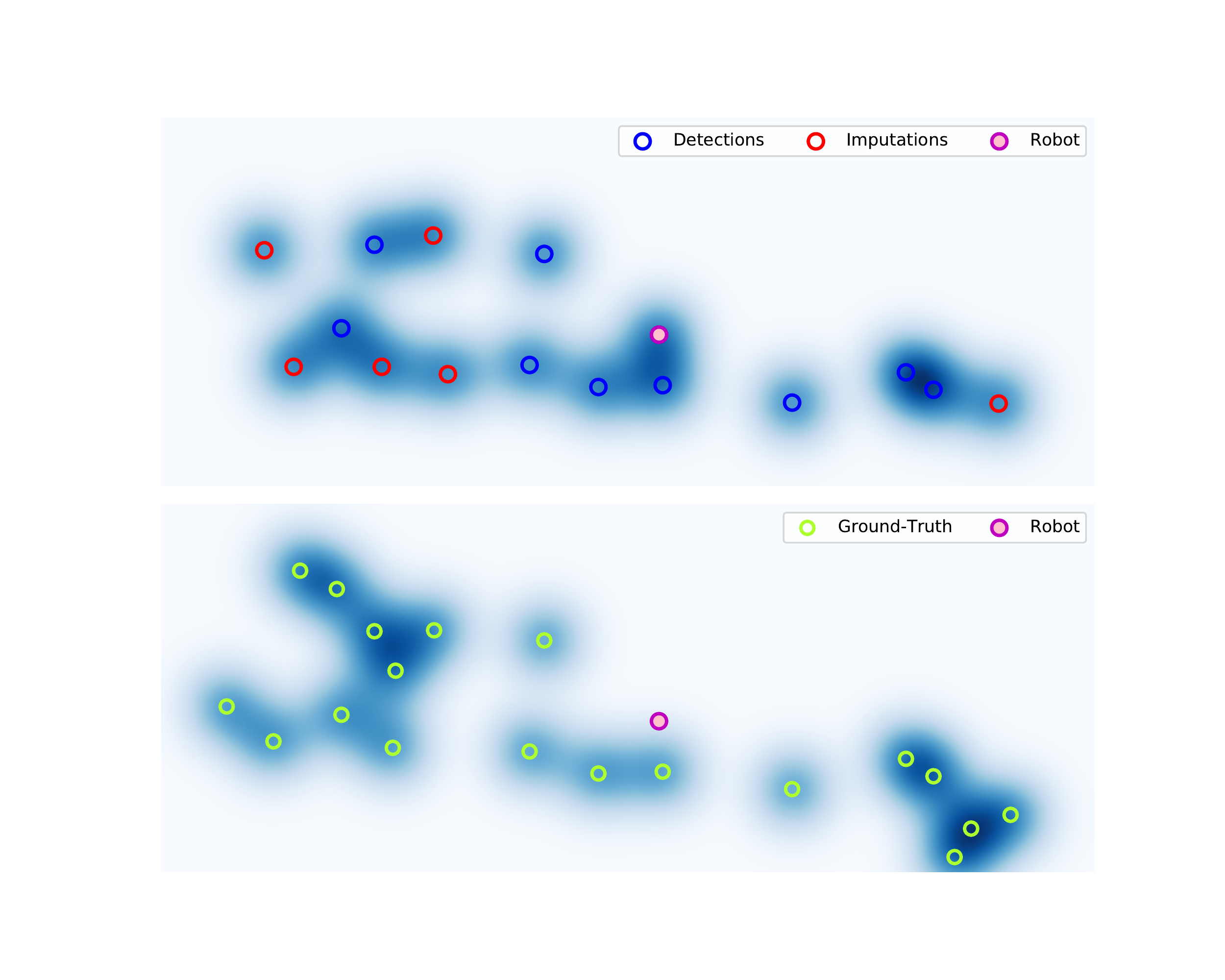}\\
			
			\includegraphics[trim=80pt 20 80pt 20, clip, width=0.42\textwidth]{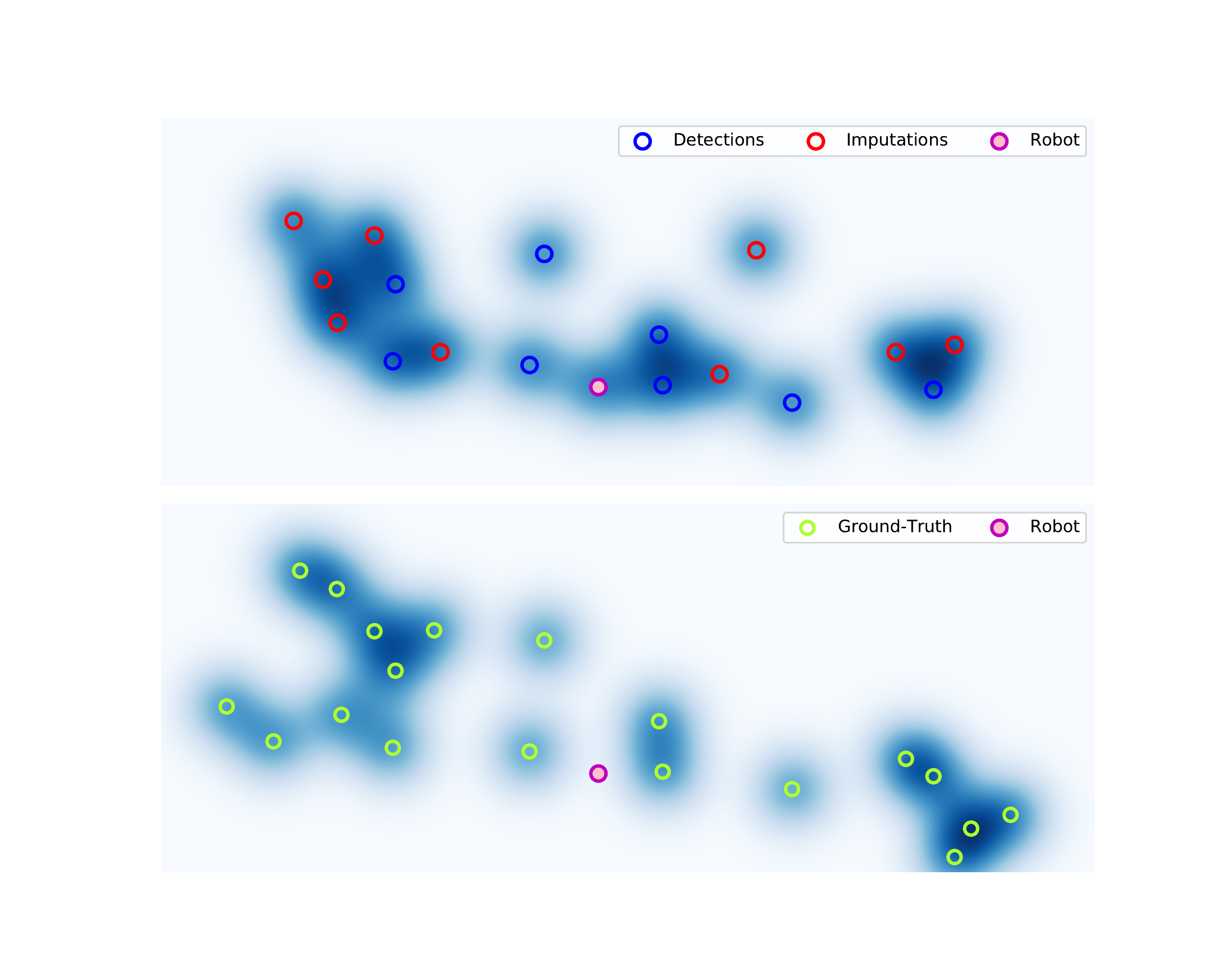}&

			\includegraphics[trim=80pt 20 80pt 20, clip, width=0.42\textwidth]{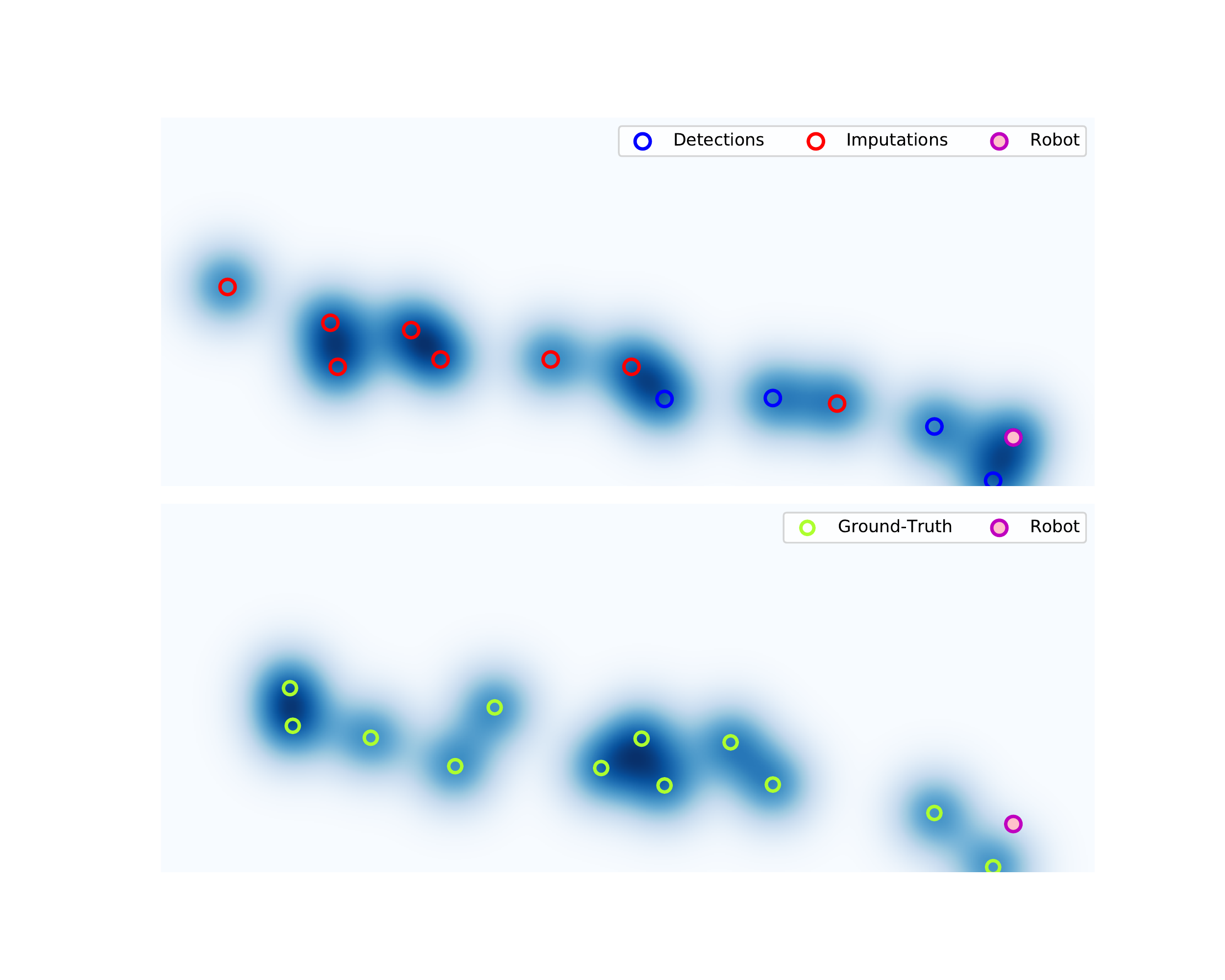}

%			\includegraphics[trim=80pt 20 80pt 20, clip, width=0.23\textwidth]{figs/new-results/104-752}& 

%			\includegraphics[trim=80pt 20 80pt 20, clip, width=0.23\textwidth]{figs/new-results/100-736}
%& 

%			\includegraphics[trim=80pt 20 80pt 20, clip, width=0.23\textwidth]{figs/new-results/111-832}
%&
			
%			\includegraphics[trim=80pt 20 80pt 20, clip, width=0.23\textwidth]{figs/new-results/112-848}
	
		\end{tabular}
	\end{center}	
	\caption{Qualitative results on the HERMES dataset. In four examples, we show the ground truth crowd (lower picture), in addition to the imputed crowd and the detections.}
	\label{fig:imputation-results}
\end{figure*}

In Fig. \ref{fig:imputation-results} some imputation examples are shown for Hermes dataset. The algorithm has proposed interesting samples in many cases using the social-tie patterns it has learned.

% !TeX root=main.tex

\section{Conclusion and Future Work}

We have proposed a new approach to impute the structure of a crowd in which a robot is performing navigation with its limited sensing. We leverage several new concepts that we have introduced to describe crowd patterns around the robot (strong and absent ties, communities and territories) to form a generative model for crowd patterns and use it to samples of imputed occupation maps, based on what is observed through the robot perception. We have shown on real crowd datasets that the proposed indicators reflect the nature of the typical pairwise relation within crowds, and we have obtained competitive prediction results, in particular on datasets with well-structured pedestrian flows.       
\label{sec:conclusion}

\bibliographystyle{acm}
\bibliography{ms}

\end{document}